\documentclass[a4paper]{article}
\pdfoutput=1

\usepackage[preprint]{spconf-arxiv}
\usepackage{graphicx,amsmath,amssymb,setspace,cite,algorithm, algorithmic}
\usepackage{epstopdf}
\usepackage{pgfplots}
\pgfplotsset{compat=1.3}
\usepackage{tikz}							
\usetikzlibrary{patterns}
\usepackage[position=t]{subfig}
\usepackage{stfloats}

\copyrightnotice{\copyright\ IEEE 2015}
                 \toappear{To appear in {\it Proc.\ EUSIPCO2015,
                 August 31- September 04, 2015, Nice, France}}

\renewcommand{\u}{\boldsymbol u}

\newcommand{\p}{\boldsymbol p}
\newcommand{\q}{\boldsymbol q}
\renewcommand{\r}{\boldsymbol r}
\newcommand{\x}{\boldsymbol x}
\newcommand{\w}{\boldsymbol w}
\newcommand{\y}{\boldsymbol y}
\newcommand{\z}{\boldsymbol z}
\newcommand{\K}{\boldsymbol K}

\renewcommand{\L}{\boldsymbol L}

\newcommand{\divk}[2]{\text{div}^{#2}[{#1}]}
\newcommand{\R}[1]{\mathbb{R}^{#1}}

\newcommand{\TGV}[1][2]{\text{TGV}_\alpha^{#1}}

\newlength{\beforeSS}
\newlength{\afterSS}
\newlength{\hspaceFig}
\newlength{\sizeFig}
\let\OLDthebibliography\thebibliography
\renewcommand\thebibliography[1]{
  \OLDthebibliography{#1}
  \setlength{\parskip}{0pt}
  \setlength{\itemsep}{0pt plus 0.1ex}
}

\pgfkeys{/pgfplots/scale/.style={
  x post scale=#1,
  y post scale=#1,
  z post scale=#1}
}

\pgfkeys{/pgfplots/axis labels at tip/.style={
    xlabel style={at={(current axis.right of origin)}, xshift=4.7ex, anchor=center},
    ylabel style={at={(current axis.above origin)}, yshift=0ex}}
}

\title{{Post-Reconstruction Deconvolution of PET Images}\\{by Total Generalized Variation Regularization}}
\name{St\'ephanie Gu\'erit$^*$, Laurent Jacques$^{*\mathsection}$, Beno\^it Macq$^*$, John A. Lee$^{*\dagger\mathsection}$%
	\thanks{$^\mathsection$ Research Associates with the Belgian F.R.S.-FNRS.}%
}
\address{
	\normalsize $^*$ ISPGroup/ICTEAM, $^\dagger$ MIRO/IREC, Universit\'e catholique de Louvain, Belgium\\
}
\begin{document}

\maketitle
\begin{abstract}

Improving the quality of positron emission tomography (PET) images, affected by low resolution and high level of noise, is a challenging task in nuclear medicine and radiotherapy. This work proposes a restoration method, achieved after tomographic reconstruction of the images and targeting clinical situations where raw data are often not accessible. Based on inverse problem methods, our contribution introduces the recently developed total generalized variation (TGV) norm to regularize PET image deconvolution. Moreover, we stabilize this procedure with additional image constraints such as positivity and photometry invariance. A criterion for updating and adjusting automatically the regularization parameter in case of Poisson noise is also presented. Experiments are conducted on both synthetic data and real patient images. 
\end{abstract}

\section{Introduction}
\label{sec:intro}
In radiotherapy, positron emission tomography (PET) images provide oncologists with useful information about the metabolic activity of the patient. The accumulation of the $^{18}$F-fluorodeoxyglucose ($^{18}$F-FDG) radiotracer in tissues of abnormally high metabolic activity, \emph{e.g.}, cancer cells, leads to the emission of positrons later detected by the PET imaging system \cite{Valk2004}. Combined with anatomical CT or MR images, these functional images are widely used for diagnosis, monitoring during the treatment and follow-up of the patient after therapy. 

A challenging step in radiotherapy treatment is the accurate delineation of the tumour volume based on PET images \cite{Lee2010a}. Some methods rely on visual interpretation or on activity threshold determination but suffer from methodological limitations \cite{Lee2010a}. The use of more complex segmentation algorithms is impeded by the two main drawbacks of PET imaging: low spatial resolution and high level of noise \cite{Valk2004}. The first one is mainly due to the blur introduced by the random positron travel, the photon diffusion in the patient tissues before annihilation and the large size of the scintillators required to detect high-energy photons. The resulting point spread function (PSF) of the PET system is generally not uniform and can vary slightly in the field of view (FOV). The latter, \emph{i.e.}, the noise in raw data, is due to low photon detection efficiency of the detectors and the limitation of the injected tracer dose for obvious radio-protective reasons. The noise is Poissonian in the projection space, before reconstruction.

Improving the quality of PET images is essential before applying more advanced segmentation techniques. Deblurring and denoising can be applied either during the reconstruction process or in a post-processing phase. The second approach is more appropriate for clinical use since the access to raw data or to the reconstruction algorithm of the scanner is not always possible. After sinogram correction and standard iterative 3D reconstruction algorithms like OSEM \cite{Tong2010}, noise in image is still multiplicative in first approximation (scaled Poisson). Nowadays, the most widely used tool in clinical denoising is a classical Gaussian filter. Edge-preserving filters such as bilateral filters or M-smoothers are also used but, like the Gaussian filter, they are generally designed for additive noise \cite{Geets2007a}. This issue can be overcome using first a variance-stabilizing transform (VST) such as proposed by Anscombe \cite{Anscombe1948}. For the deblurring step, classical deconvolution algorithms like Van Cittert's or Landweber's \cite{Geets2007a} are used. Other methods combine the denoising and deblurring steps by solving an optimization problem (inverse problem) \cite{Getreuer2012a,Dupe2012}. 

In this paper, we choose the inverse problem approach in a post-processing phase and propose a problem formulation specific to the restoration of PET images. Our method introduces the total generalized variation (TGV) as a regularization term \cite{Bredies2010} for PET deconvolution, considers specific properties of PET physics such as positivity and photometry invariance and takes into account the Poisson statistics of noise through the data fidelity term and the choice of the regularization parameter $\lambda$. A criterion for automatic selection of $\lambda$ is presented. The validation of the TGV algorithm is made on synthetic images and on real medical images where the PSF is well estimated.

The paper is organized as follows. Sec.~\ref{sec:theory_impl} introduces the mathematical model and the underlying assumptions, as well as theoretical and numerical aspects related to the convex optimization problem. The experimental setup is described in Sec.~\ref{sec:experiments}, along with the results, which are discussed in Sec.~\ref{sec:discussion}. Conclusions are presented in Sec.~\ref{sec:conclusion}.

\section{Forward model and inverse method}
\label{sec:theory_impl}
	In this section, we describe the forward model and the assumptions we made. We then explain the formulation of the inverse problem based on a maximum \textit{a posteriori} estimator (MAP). Finally, we give some details about the numerical implementation and the choice of the parameters

\subsection{Forward model}
\label{sec:forward}
	
For the sake of simplicity, we work with a discretized model. The two dimensional images of size $N_1\times N_2$ belong to the Euclidean space $\R{N}$ with $N = N_1N_2$. Let $\u_0 \in \R{N}$ be the original and unkown PET image of the functional processes in a patient body. After tomographic reconstruction, the observed image $\z \in \R{N}$ is associated with $\u_0$ through the forward model
	\begin{equation*}
		\z = \mathcal{N}_\nu(\K(\u_0)),
		\label{eq_modele}
	\end{equation*}
	where $\K$ is a blur operator accounting for both the physical and instrumental inaccuracies and $\mathcal N_\nu$ is a noise corruption of parameter $\nu$. Some assumptions on $\K$ and $\mathcal N_\nu$ lead to a simpler model. In first approximation, the PSF near the center of the FOV is uniform, Gaussian and isotropic in the 2D-plane \cite{Geets2007a}. Consequently, the blurred image is assumed to result from the convolution of the original image with the PSF of the scanner. As mentionned in Sec.~\ref{sec:intro}, noise is considered as Poissonian in first approximation. 
	A simpler model is 
	\begin{equation}
		\z = \mathcal{P}(\K\u_0), 
		\label{eq_modele_general}
	\end{equation}
	where the convolution operator $\K \in \R{N\times N}$ is assumed to be known (see Sec.~\ref{subsubsec:exp_method}), linear (uniform PSF) and bounded. Each pixel $i$ of the original image is corrupted by Poisson noise $\mathcal{P}$ with mean $(\K\u_0)_i$.
	
	Physics of PET imaging suggests two additional constraints to model (\ref{eq_modele_general}). 	
	The first property is the \textit{positivity} since the original image $\u_0$ represents a nonnegative activity in a phantom or \textit{in vivo}, \emph{e.g.}, the metabolic activity. Hence,
				\begin{equation*} 
				\u_0 \succeq \boldsymbol{0}.
				\label{positivity}
				\end{equation*} 					
	The second property is \textit{photometry invariance}. Total photometry preservation means that the total photon counts in the original and observed images are approximately the same. This property is particularly valuable for quantification aspects like the measurement of standardized uptake values \cite{Valk2004},
				\begin{equation*}  
				\sum_{i=1}^{N} (\u_0)_i \approx  \sum_{i=1}^{N} z_i.
				\label{photometry}
				\end{equation*}
	
\subsection{Inverse problem formulation}
		
Solving an inverse problem consists in finding an estimator $\hat\u_0$ of the original image $\u_0$ from observations $\z$ in (\ref{eq_modele_general}). Noise and the low-pass filter effect of the PSF lead to an ill-posed problem for which direct inversion is not possible and unicity of the solution is not guaranteed. 	To address this issue and reduce the set of possible solutions, a widely used method is to regularize the problem \cite{Boyd2004}. A penality term encourages the solution to respect a certain prior model of the original image.
		
	\subsubsection{\textit{A priori} knowledge on the PET images}
	\label{sec:apriori}
				
	A possible assumption about the image concerns its \textit{piecewise smoothness}. In many applications, frameworks based  on total variation (TV) are used to promote piecewise constant images \cite{Bredies2010}. However, as Knoll et al. \cite{Knoll2011} mentionned in the case of MRI, this type of regularization is not well adapted to real medical images. Since these images are not piecewise constant, staircasing artifacts are observed in smooth image areas with TV regularization \cite{Bredies2010,Knoll2011}. The total generalized variation (TGV), introduced by Bredies et al. \cite{Bredies2010} can be considered as the generalization of TV to higher-order image derivatives. Simplifying its presentation to our conventions, the second order TGV of $\x\in\R{N}$ is defined in a discrete setting as 
				\begin{equation}
				\TGV(\x) = \min_{\w\in \R{N\times 2}}~\|\nabla \x - \w \|_{2,1} + \alpha \|\varepsilon(\w)\|_{2,1},
				\label{eq_TGV_ordre2}
				\end{equation}
				where $\alpha\in\mathbb{R}$ is a positive constant balancing between the edge-preserving term and the smoothness-promoting term. The general discrete gradient operator $\nabla$ applicable to $N\times k$ tensor fields is defined as
				\begin{equation*}
					\nabla : \R{N\times k} \to \R{N\times 2k},~ \x \mapsto (\nabla \x) = (\nabla_1 \x, \nabla_2 \x),
				\end{equation*}
				with $k\in\mathbb{N}_0$ and $\nabla_i \in \R{N\times N}$ the first spatial derivative of the tensor field in direction $\boldsymbol e_i$. The symmetrized derivative operator $\varepsilon$ is 
				\begin{equation*}
					  \varepsilon : \R{N\times 2} \to \R{N\times 4},~ \x \mapsto \varepsilon(\x) = \frac{1}{2}\left((\nabla\x) + (\nabla \x) S_{23}\right),
				\end{equation*}
				where $S_{23} \in \{0,1\}^{4\times 4}$ is a matrix permuting the second and the third columns of $(\nabla \x)$. Applying $\varepsilon$ on the gradient of an image provides information about its second derivative. The concept of symmetrized tensors is detailed in \cite{Bredies2010}. Finally, the $L_{p,q}$ norm of $\x \in \R{N\times k}$ is defined as 
				$$  \|\x\|_{p,q} = \left(\sum_{i=1}^N\|\x_i\|_p^q \right)^{\frac{1}{q}}.$$
				Deriving $\TGV(\x)$ is not as easy as TV$(\x)$ because an additional minimization problem has to be solved. The optimal value of $\w$ in (\ref{eq_TGV_ordre2}) depends on the features of image $\x$. Locally, in smooth areas, $\w$ is close to $\nabla \x$ to give more importance to the smoothness-promoting term. In regions near image edges, $\w$ is close to $\boldsymbol 0$ to preserve sharp edges like TV. These properties make $\TGV$ more adapted to real PET medical images than TV. Moreover, the absence of staircasing artifacts improves the efficiency of subsequent segmentation algorithms. 
				
	\subsubsection{Bayesian approach}
	To take into account this \textit{a priori} knwoledge on $\u_0$, the best estimator choice is the Bayesian MAP estimator, defined as	
				\begin{eqnarray}
				\nonumber
				\hat \u_0		& = & \underset{\u \in \R{N}}{\text{arg~max}}~ p(\u|\z) \\
				\label{eq_bayesian}
							& = & \underset{\u \in \R{N}}{\text{arg~min}}~ {-\log{p(\z|\u)}} -\log{p(\u)}.
				\end{eqnarray}
				The first term of (\ref{eq_bayesian}) is the fidelity term. This term encourages the estimated image to be close to observations $\z$ and depends directly on the noise statistics. In the case of PET images, it is the negative log-likelihood of the Poisson distribution, \emph{i.e.}, 
				\begin{equation*}
				  -\log{p(\z|\u)} = \sum_{i=1}^N (\K\u - \z \cdot f(\K\u))_i + r(\z),
				\label{data_term}
				\end{equation*}
				where $r(\z)$ depends only on $\z$, $f$ is applied componentwise on vector with $f(t) = \log{t}$ if $t>0$ and 0 otherwise and $\cdot$ denotes the Hadamard (or elementwise) product.
				The second term in (\ref{eq_bayesian}) is the regularization term promoting the respect of the \textit{a priori} knowledge, \emph{i.e.}, a minimal $\TGV$. Indicator functions $\imath_{\mathbb{R}_+^N}$ and $\imath_{\mathcal{C}}$ take into account positivity and photometry invariance constraints. They equal zero if the constraint is respected and $+\infty$ otherwise.

				The nonsmooth convex optimization problem reads
		\begin{equation} 
			\underset{\u \in \R{N},\atop \w \in \R{N\times 2}}{\text{min}}\quad \lambda~\sum_{i=1}^N (\K\u - \z \cdot f(\K\u))_i  + \|\nabla \u - \w \|_{2,1} + \alpha \|\varepsilon(\w)\|_{2,1}  + \imath_{\mathbb{R}_+^N}(\u) + \imath_{\mathcal{C}}(\u),
			\label{eq_minimisation}
		\end{equation}
	where $\lambda > 0$ is the regularization parameter, \emph{i.e.}, the trade-off between the data fidelity and  regularization terms. The value $\alpha = 2$ is used in practice \cite{Knoll2011}.

	\subsection{Numerical implementation}
		The discrete setting is described in details in \cite{Bredies2010}. Discrete operators $\nabla_1$ and $\nabla_2$ are approximated using forward and backward first-order finite differences. The convolution of the image with operator $\K$ is made with a FFT. The undesired border effects are avoided using a ``padding'' method presented by Almeida et al. \cite{Almeida2013}. 
		
\subsubsection{Chambolle-Pock primal-dual algorithm}
Let $\L: \mathcal X \rightarrow \mathcal Y$ be a linear continuous operator with a norm defined by $\|\L\|_2 = \max\{\|\L \x\|_2~|~\x \in \mathcal X \text{ with } \|\x\|_2 \le 1\} $ and $G: \mathcal X \rightarrow {[0,{+\infty}[}$ and $F^\star: \mathcal Y \rightarrow {[0,{+\infty}[}$ be proper, convex, and lower-semicontinuous functions.
The Chambolle-Pock (CP) primal-dual algorithm \cite{Chambolle2010} is designed to solve the following saddle-point problem 
			\begin{equation}
				\begin{aligned}
					& \underset{\x \in \mathcal X}{\text{min}}~\underset{\y \in \mathcal Y}{\text{max}}
	 				& & \langle \L \x,\y \rangle - F^\star(\y) + G(\x),
				\end{aligned}
				\label{eq_minmax_chambolle}
			\end{equation}
which is a primal-dual formulation of the primal problem $\underset{\x \in \mathcal X}{\text{min}}\ \ F(\L \x) + G(\x)$.

The CP algorithm belongs to the family of proximal algorithms \cite{Parikh2013}. Such algorithms are based on the notion of proximal operator and can deal with optimization of non differentiable functions. Let $\varphi$ be a lower semicontinuous convex function from $\mathcal X$ to $\left]-\infty,+\infty\right[$ such that $\text{dom}f$ is non empty. The proximal operator of $\varphi : \mathcal X \rightarrow \mathcal X$ evaluated in $\tilde \x \in \mathcal X$ is unique and defined as \cite{Parikh2013} 
\begin{equation}  
	\text{prox}_{\varphi}(\tilde \x) := \underset{\x \in \mathcal X}{\text{arg~min}}~\frac{1}{2}\|\tilde \x - \x\|^2_2 + \varphi(\x).
\label{prox_op}
\end{equation}

In our case, the presence of the two indicator functions leads to a nonsmooth and non differentiable objective function. The use of a proximal algorithm and particularly CP is appropriate \cite{Anthoine2012}. A compact formulation of (\ref{eq_minimisation}) is given by
\begin{equation*}
\underset{\u \in \R{N},\atop \w \in \R{N\times 2}}{\text{min}}\ \ F_1(\K\u) + F_2(\nabla \u - \w) + F_3(\varepsilon(\w)) + G(\u),
			\label{eq_min_compact}
\end{equation*}
with straightforward definitions of $F_1$, $F_2$, $F_3$ and $G$ from (\ref{eq_minimisation}).

A primal-dual formulation similar to (\ref{eq_minmax_chambolle}) of this problem is derived in a product space \cite{Gonzalez2014} using duality principles \cite{Chambolle2010,Boyd2004} and leads to Algorithm \ref{algo_PD_PET}, \emph{i.e.}, the CP algorithm adapted to TGV denoising and deblurring of PET images. Function $\varphi^\star$ is the Legendre-Fenchel conjugate of any function $\varphi$. The divergence operator $\text{div}$ is defined as equal to $-\nabla^*$ with
\begin{equation*}
	\nabla^* : \R{N\times 2k} \to \R{N\times k},~ \x = (\x_1,\x_2) \mapsto (\nabla_1^* \x_1 + \nabla_2^* \x_2).
\end{equation*}

\subsubsection{Proximal operators}

Let $\tilde \x \in \R{N}$. The proximal operator of the primal function $G$ is given by a combination of positivity (max) and photometry invariance (average) proximal operators \cite{Parikh2013},
\begin{equation*}  
\text{prox}_{\sigma G}(\tilde \x) = \max(\tilde \x, 0) - \frac{1}{N} \sum_{i=1}^N \left(\max(\tilde \x, 0)) - \z\right)_i.
\end{equation*}

From definition (\ref{prox_op}) and the Moreau decomposition property \cite{Parikh2013}, the proximal operator of $F_1^\star$ is, for each component $i$ of $\tilde \x \in \R{N}$ \cite{Anthoine2012},
\begin{equation*}
\text{prox}_{\sigma F_1^\star}(\tilde x_i) = \frac{1}{2}\left(\tilde x_i + \lambda - \sqrt{(\tilde x_i - \lambda)^2+ 4\sigma\lambda z_i}\right),
\end{equation*}
if $z_i$ is non-zero and is equal to parameter $\lambda$ otherwise.

By conjugation of $F_2$ and $F_3$, $F_2^\star$ and $F_3^\star$ are indicator functions. Their proximal operators are projection operators onto the convex sets $Q = \{\x \in \R{N\times 2} | \|\x\|_{2,\infty} \le 1\}$ and $R = \{\x \in \R{N\times 4} | \|\x\|_{2,\infty} \le \alpha\}$.

\algsetup{indent=1em}
\begin{algorithm}[t!] 
\small
	\caption{\small for TGV denoising and deblurring of PET images.}               
	\begin{algorithmic}[1]
		\STATE \textbf{initialization:}  $n = 0$ ; $\u^{(0)}=\bar{\u}^{(0)} = \y \in \R{N}$ ; $\w^{(0)}=\bar{\w}^{(0)} = 0 \in \R{N\times 2}$ ; $\p^{(0)} = 0 \in\R{N}$ ; $\q^{(0)} = 0 \in \R{N\times 2}$ ; $\r^{(0)} = 0 \in \R{N\times 4}$ ; choose $\tau^{(0)} = \sigma^{(0)} = 0.9 / \|\L\|_2$.
    				\REPEAT
    					\STATE $\p^{(n+1)} = \text{prox}_{\sigma^{(n)} F_1^\star}(\p^{(n)} + \sigma^{(n)} \K \bar{\u}^{(n)})$
    					\STATE $\q^{(n+1)} = \text{prox}_{\sigma^{(n)} F_2^\star}(\q^{(n)} + \sigma^{(n)}(\nabla \bar{\u}^{(n)} - \bar{\w}^{(n)}))$
    					\STATE $\r^{(n+1)} = \text{prox}_{\sigma^{(n)} F_3^\star}(\r^{(n)} + \sigma^{(n)}\varepsilon(\bar{\w}^{(n)}))$	
    					\STATE $\u^{(n+1)} = \text{prox}_{\tau^{(n)} G} (\u^{(n)} + \tau^{(n)}(\divk{\q^{(n)}}{} - \K^* \p^{(n)}))$
    					\STATE $\w^{(n+1)} = \w^{(n)} - \tau^{(n)}(-\q^{(n)} - \divk{\r^{(n)}}{})$
    					\STATE $\bar{\u}^{(n+1)} = 2\u^{(n+1)} - \u^{(n)}$
    					\STATE $\bar{\w}^{(n+1)} = 2\w^{(n+1)} - \w^{(n)}$
   				\UNTIL convergence of $\u$
	\end{algorithmic}
	\label{algo_PD_PET}
\end{algorithm}

\subsubsection{Parameters choice} 
Step-sizes $\sigma$ and $\tau$ are updated at each iteration depending on the size of the primal and dual residuals \cite{Gonzalez2014}. To ensure the convergence, condition $\sigma\tau\|\L\|^2_2 < 1$ has to be verified \cite{Chambolle2010}.

The choice of the regularization parameter is based on the discrepancy principle for Poisson noise adapted in \cite{Carlavan2011} to images with null background. Parameter $\lambda$ is selected such that
\begin{equation} 
	\text{KL}(\z,\K\hat\u_\lambda) \approx \frac{M}{2},
	\label{discrep_lambda}
\end{equation}
where $M\le N$ is the number of non zero pixels and $\text{KL}$ is the Kullback-Leibler divergence defined as
$$  \text{KL}(\x,\y) = \sum_{i=1}^N (\y - \x + \x \cdot f(\x) - \x \cdot f(\y))_i.$$
Interestingly, $\text{KL}(\z,\K\hat\u_\lambda) = F_1(\K\u) - F_1(\z)$, so that (\ref{discrep_lambda}) can be interpreted as a distance in the range of $F_1 \circ \K$.

\section{Experiments}
\label{sec:experiments}
	\begin{figure*}[t!]		
	\vspace{-1mm}
	\captionsetup[subfigure]{labelformat=empty}
	\centering
	\hspace{-11.8mm}
	\subfloat{
	\begin{tikzpicture}
		\begin{axis}[
	font = \footnotesize,
	axis line style = white,
	height = 2.5cm,
	width = 2mm,
	scale = 1,
	axis lines = left,
	xmin = -0.2, xmax = 0, 
	ymin = 0, ymax = 280,
	scale only axis, 
	xtick = \empty,
	every x tick/.style={color=white},
	every y tick/.style={color=white},
	every axis/.append style={font=\tiny},
	every axis legend/.append style={at={(1,0.3)},anchor=south east}, 
 	ytick = {0,50,...,250,260},
	yticklabels = {0,50,...,250,\hphantom{1000}},
	]
		\end{axis} 
	\end{tikzpicture}
	} \hspace{-4mm}
	\subfloat{\includegraphics[height=\sizeFig]{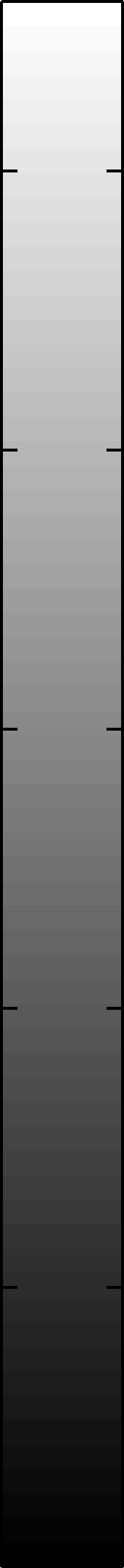}} \hspace{1mm}
	\subfloat[Original]{\includegraphics[height=\sizeFig]{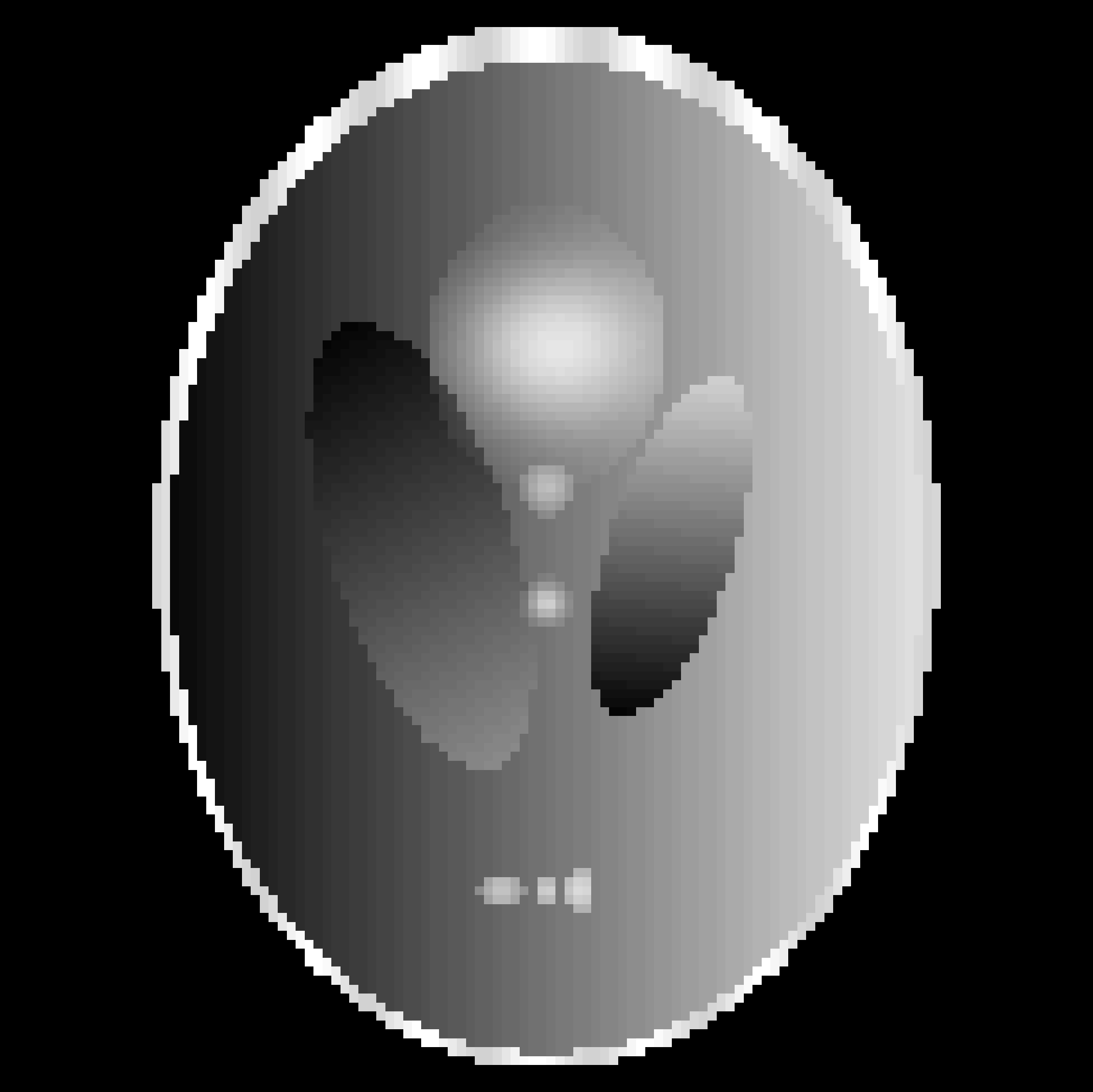}} \hspace{\hspaceFig}
	\subfloat[Corrupted]{\includegraphics[height=\sizeFig]{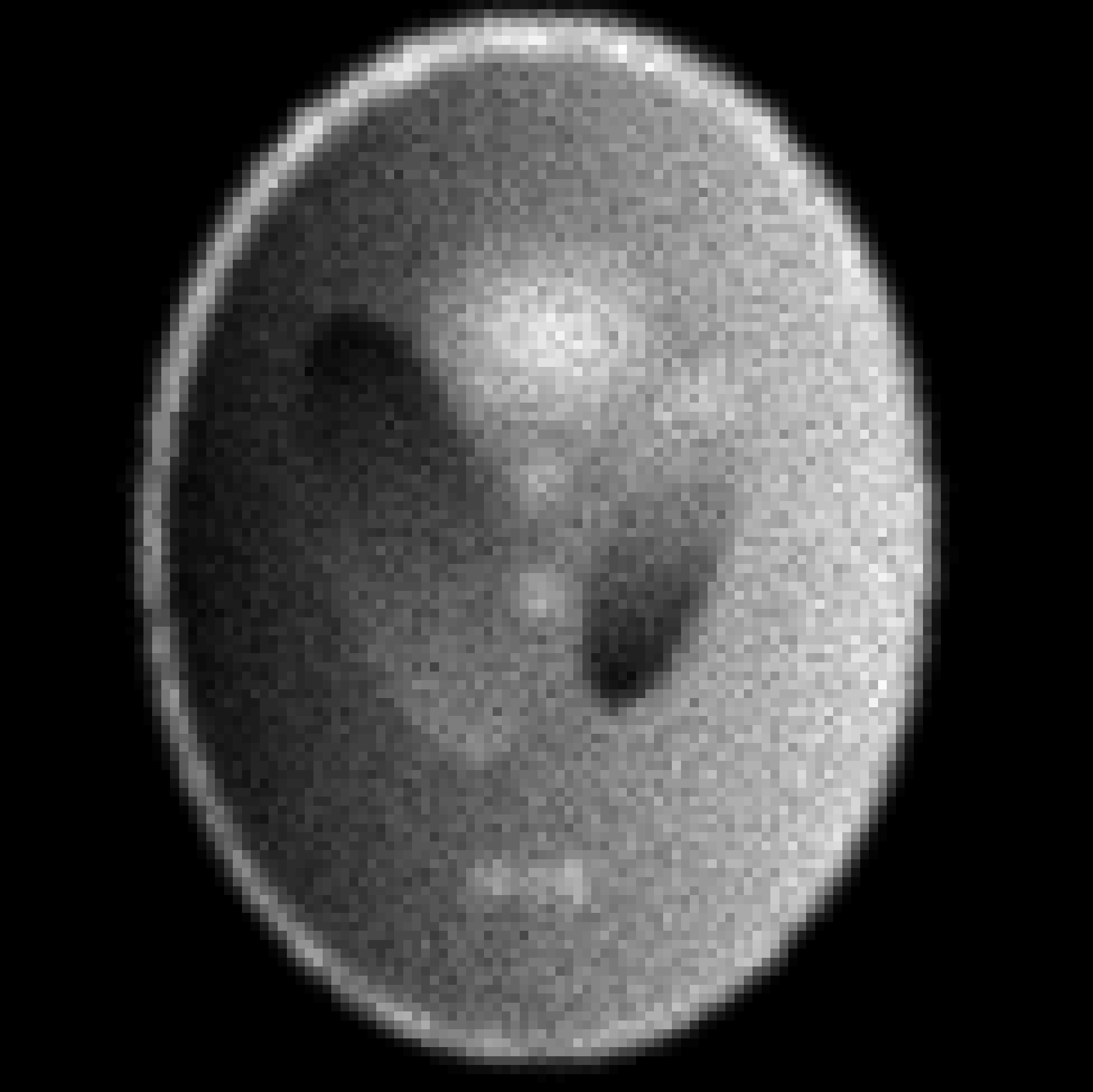}} \hspace{\hspaceFig}
	\subfloat[TV restoration]{\includegraphics[height=\sizeFig]{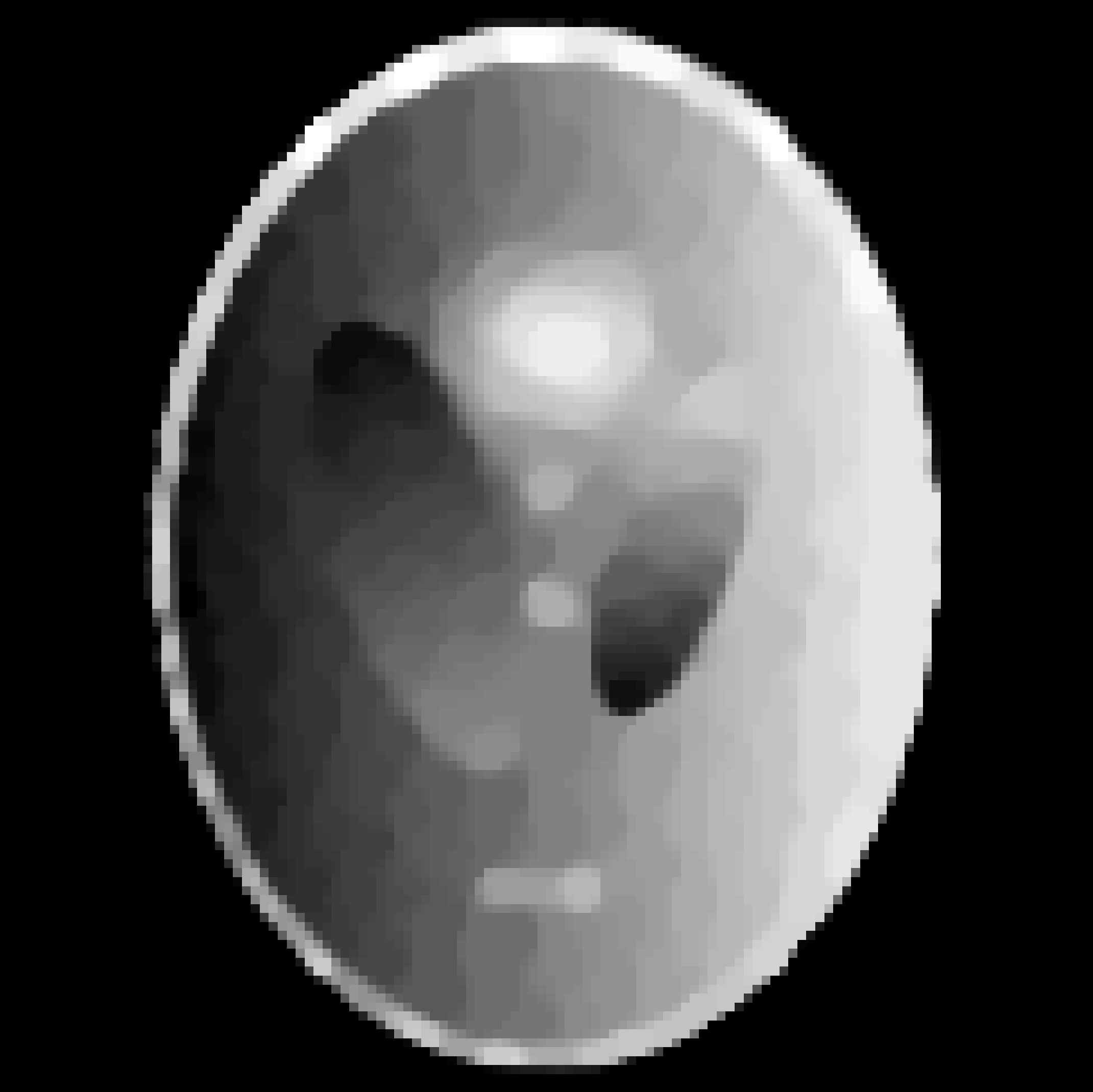}} \hspace{\hspaceFig}
	\subfloat[TGV restoration]{\includegraphics[height=\sizeFig]{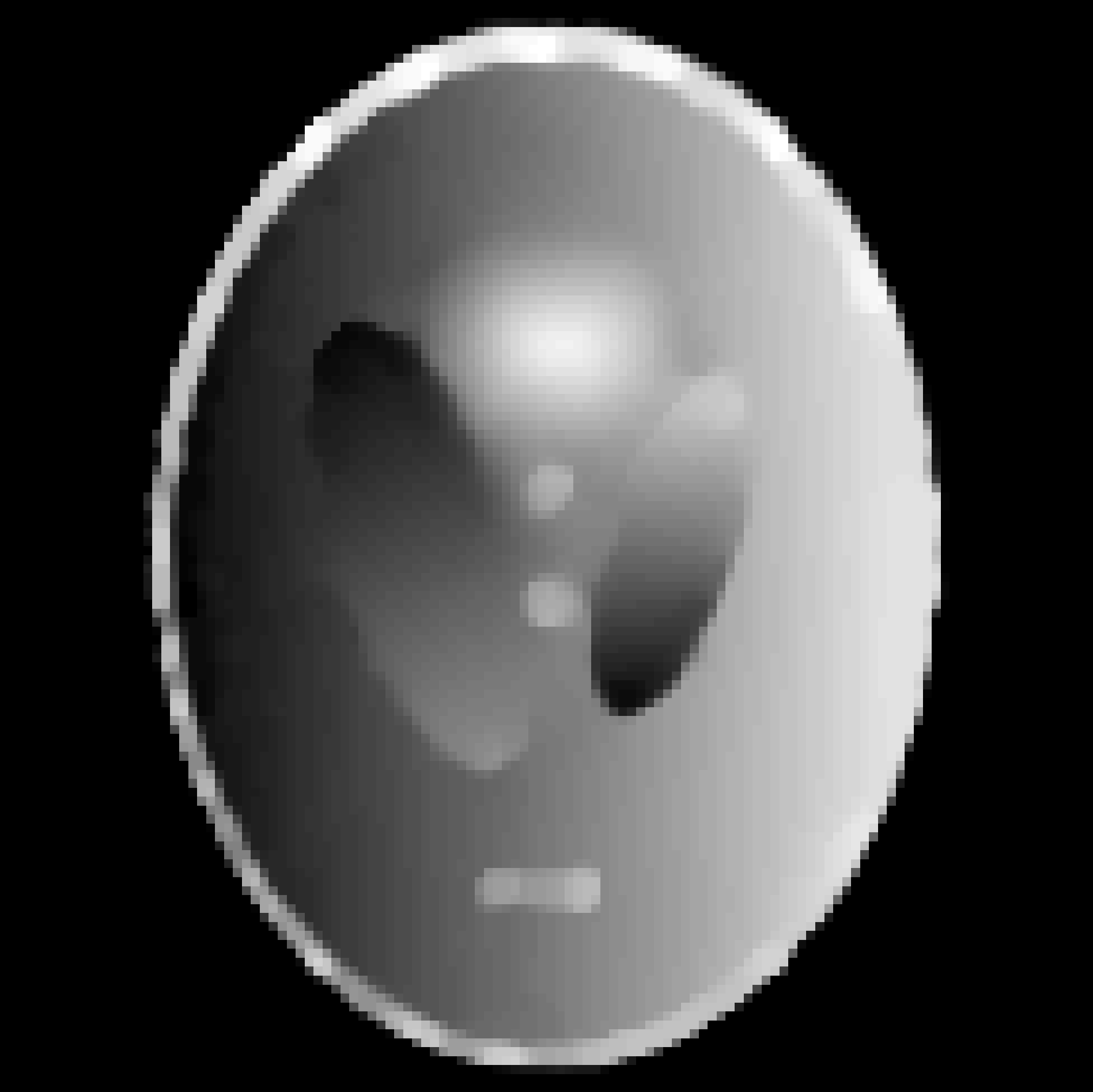}} \hspace{\hspaceFig}
	\subfloat[TGV residual]{\includegraphics[height=\sizeFig]{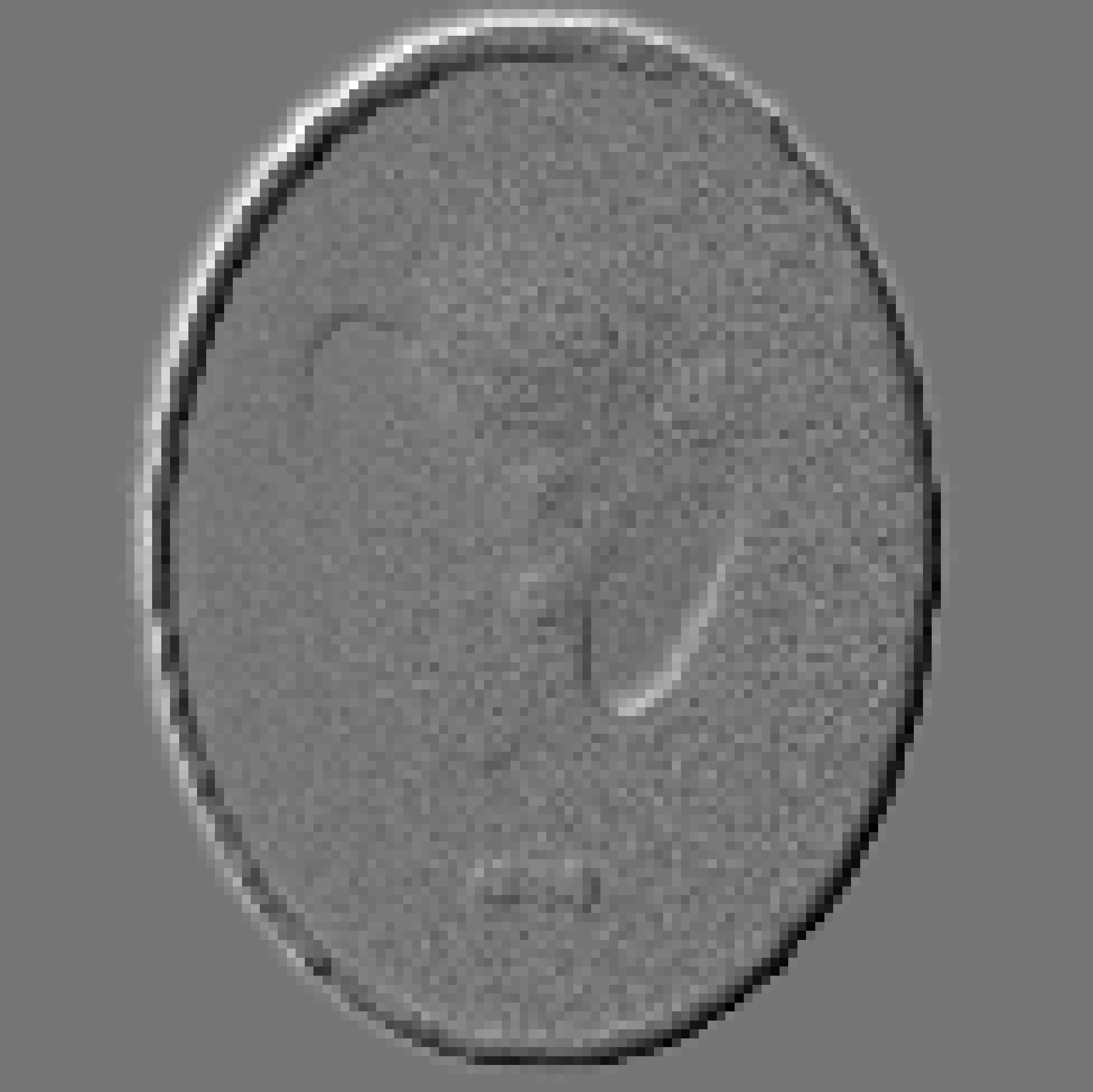}} \hspace{\hspaceFig}
	\subfloat{
	\begin{tikzpicture}
		\begin{axis}[
	font = \footnotesize,
	axis line style = white,
	height = 2.5cm,
	width = 2mm,
	scale = 1,
	axis lines = right,
	xmin = 0, xmax = 0.2, 
	ymin = -200, ymax = 200,
	scale only axis, 
	xtick = \empty,
	every x tick/.style={color=white},
	every y tick/.style={color=white},
	every axis/.append style={font=\tiny},
	every axis legend/.append style={at={(1,0.3)},anchor=south east}, 
 	ytick = {-150,-100,0,...,200},
	yticklabels = {\hphantom{-1000},-100,0,100,200},
	]
		\end{axis} 
	\end{tikzpicture}
	} \hspace{-15mm}
	\subfloat{\includegraphics[height=\sizeFig]{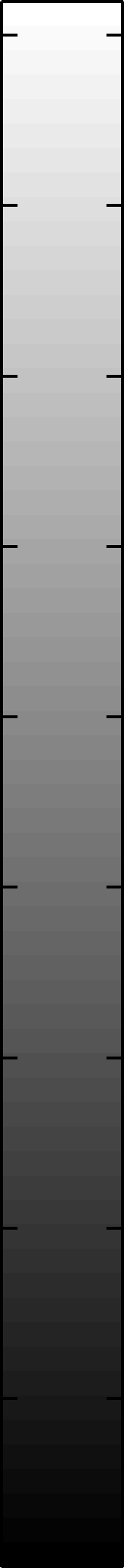}}\\[0.5mm]
	\subfloat{
	\begin{tikzpicture}
		\begin{axis}[
	font = \footnotesize,
	axis line style = white,
	height = 2.5cm,
	width = 2mm,
	scale = 1,
	axis lines = left,
	xmin = -0.2, xmax = 0, 
	ymin = 0, ymax = 28,
	scale only axis, 
	xtick = \empty,
	every x tick/.style={color=white},
	every y tick/.style={color=white},
	every axis/.append style={font=\tiny},
	every axis legend/.append style={at={(1,0.3)},anchor=south east}, 
 	ytick = {0,5,...,25,27},
	yticklabels = {0,5,...,25,\hphantom{1000}},
	]
		\end{axis} 
	\end{tikzpicture}
	} \hspace{-4mm}
	\subfloat{\includegraphics[height=\sizeFig]{figures/colorbar_original_1.png}} \hspace{1mm}
	\subfloat{\includegraphics[height=\sizeFig]{figures/phantom_ori_anoise12.png}} \hspace{\hspaceFig}
	\subfloat{\includegraphics[height=\sizeFig]{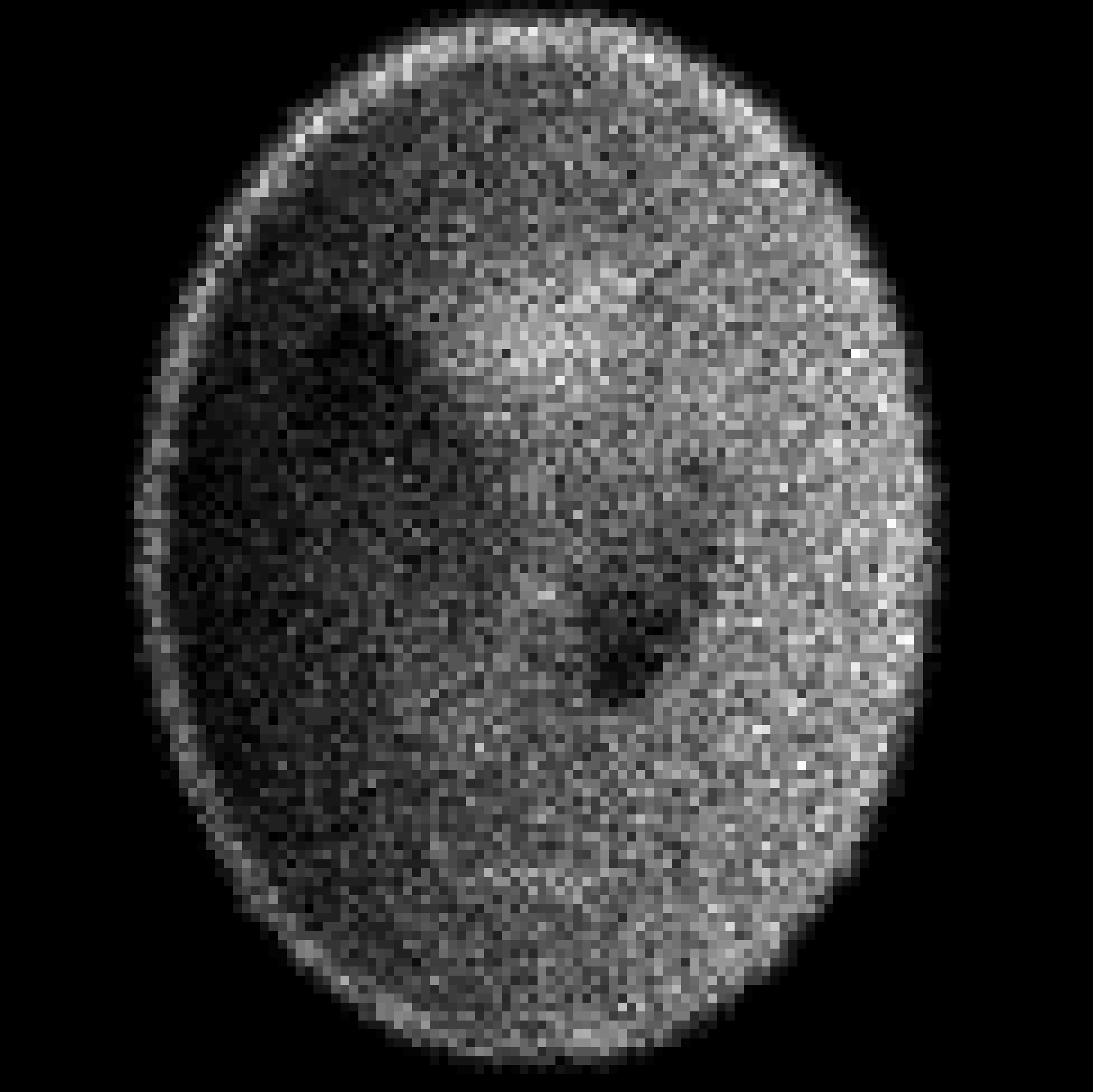}} \hspace{\hspaceFig}
	\subfloat{\includegraphics[height=\sizeFig]{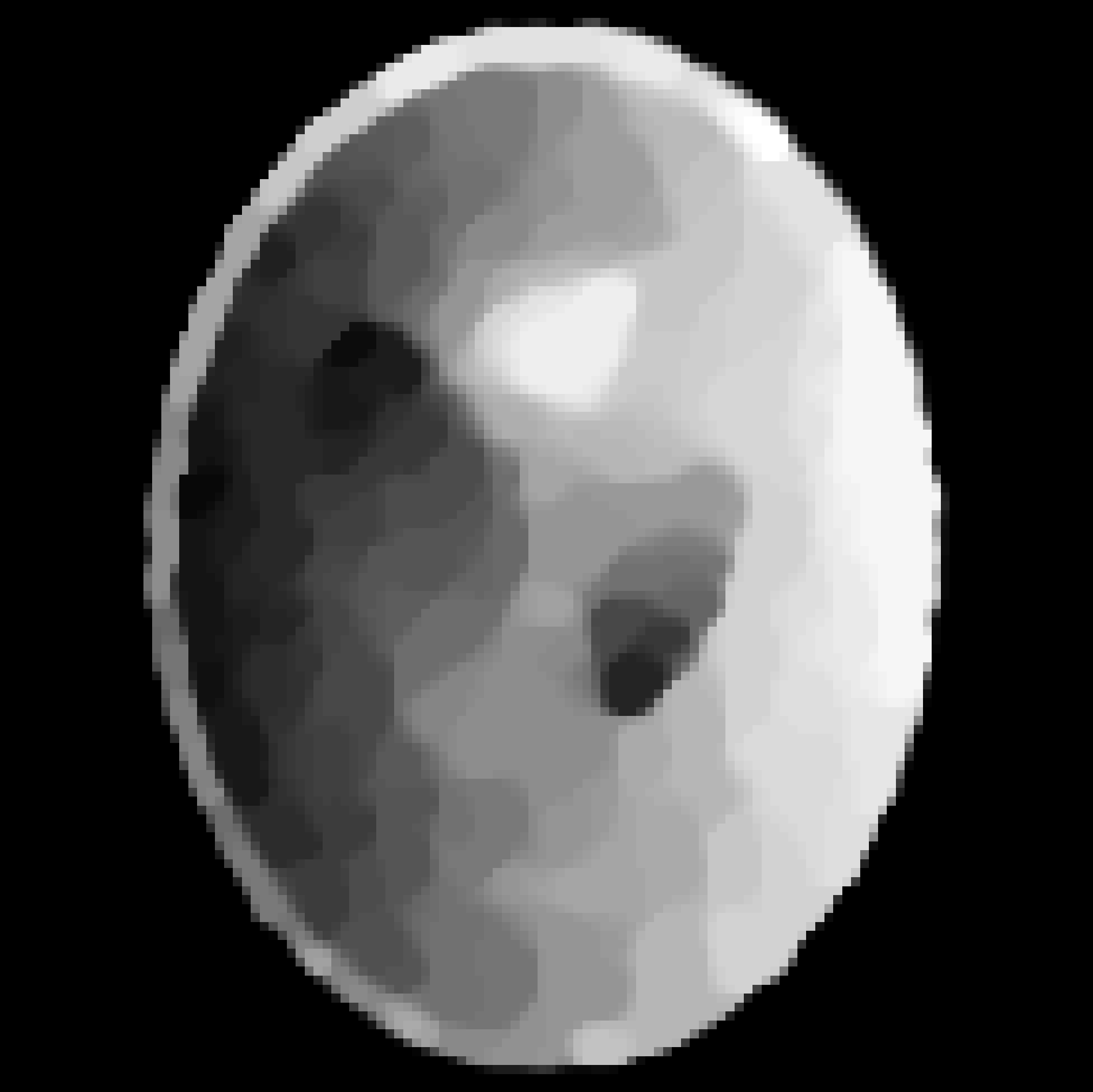}} \hspace{\hspaceFig}
	\subfloat{\includegraphics[height=\sizeFig]{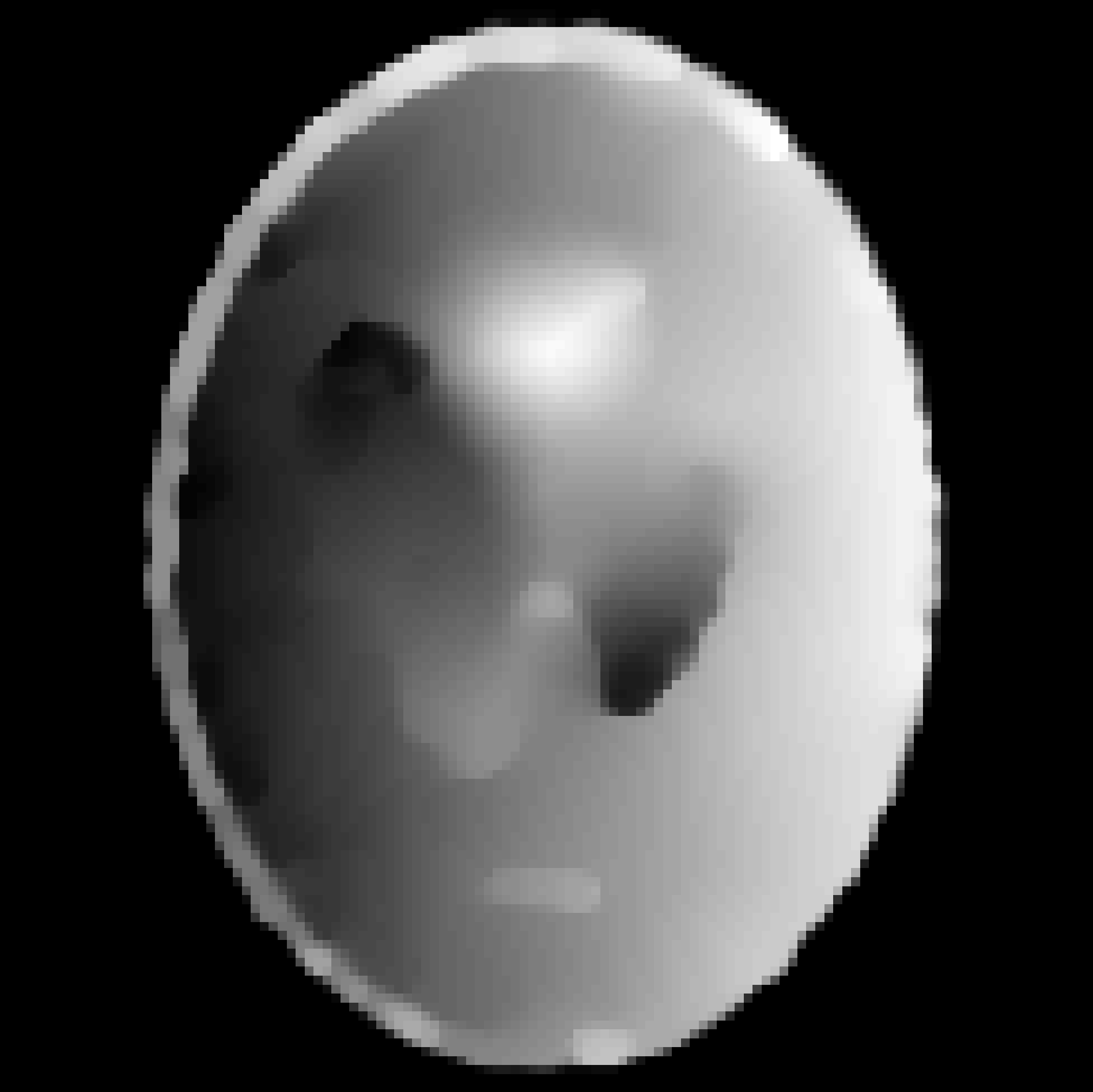}} \hspace{\hspaceFig}
	\subfloat{\includegraphics[height=\sizeFig]{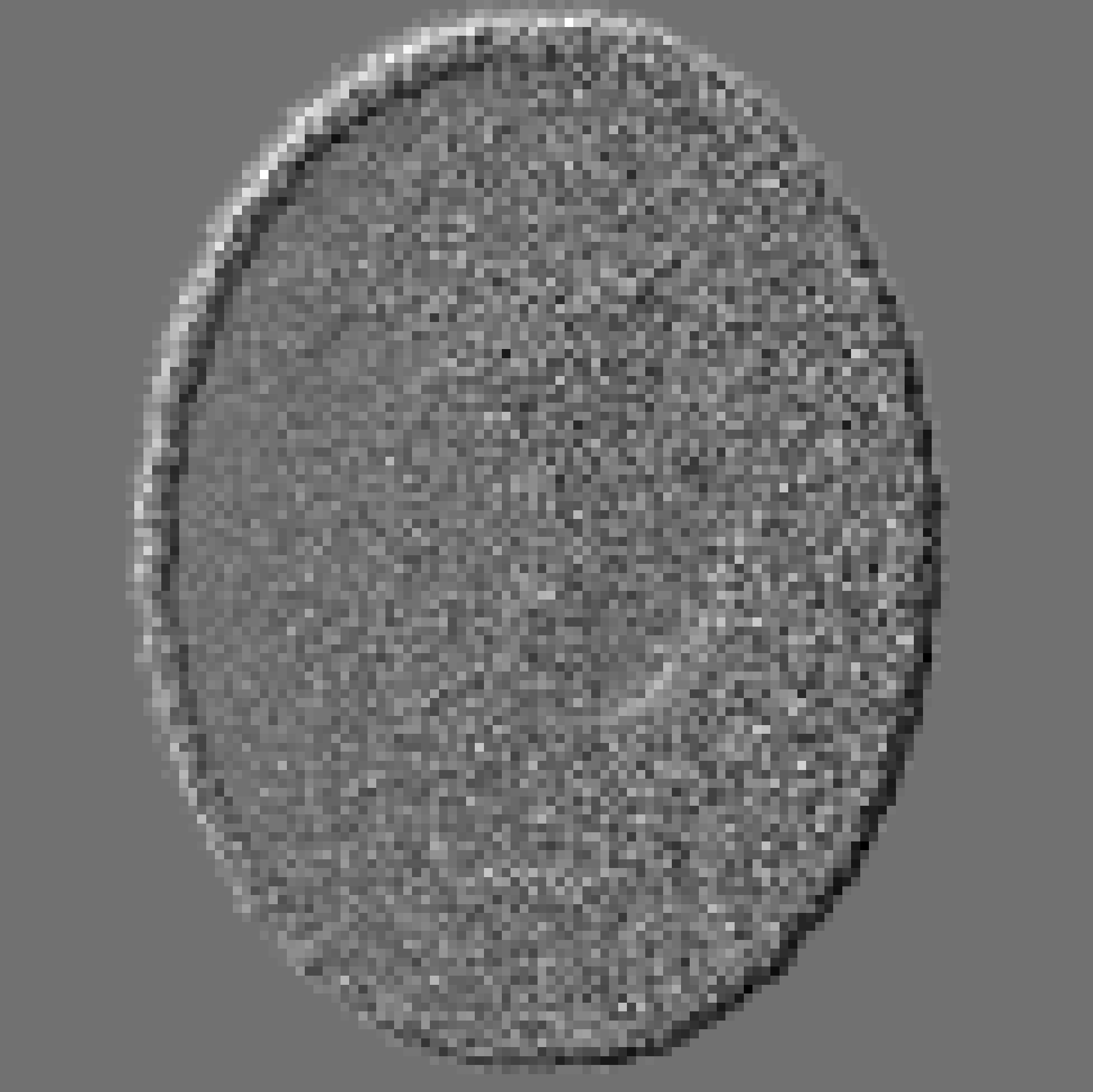}} \hspace{\hspaceFig}
	\subfloat{
	\begin{tikzpicture}
		\begin{axis}[
	font = \footnotesize,
	axis line style = white,
	height = 2.5cm,
	width = 2mm,
	scale = 1,
	axis lines = right,
	xmin = 0, xmax = 0.2,
	ymin = -20, ymax = 26,
	scale only axis, 
	xtick = \empty,
	every x tick/.style={color=white},
	every y tick/.style={color=white},
	every axis/.append style={font=\tiny},
	every axis legend/.append style={at={(1,0.3)},anchor=south east}, 
 	ytick = {-20,-10,...,20,25},
	yticklabels = {-20,-10,...,20,\hphantom{-1000}},
	]
		\end{axis} 
	\end{tikzpicture}
	}\hspace{-15mm}
	\subfloat{\includegraphics[height=\sizeFig]{figures/colorbar_residual_1.png}} \hspace{10mm}
	\vspace{-2mm}
	\caption{\small Results on synthetic data. From left to right: original image, corrupted images (convolution with isotropic Gaussian kernel and Poisson noise: $\beta_{\text{top}} = 1$ and $\beta_{\text{bottom}} = 0.1$), TV restored image ($\lambda_{\text{top}} = 32.8$, $\lambda_{\text{bottom}}=4.7$), TGV restored image ($\lambda_{\text{top}} = 25.3$, $\lambda_{\text{bottom}}=5.6$) and residual image computed as the difference between the original and the TGV restored images.}
	\label{figures_SL}
\end{figure*}
\textbf{Synthetic images.}
The synthetic image used for a first validation of the method is a modified Shepp-Logan phantom with intensities in $[0,255]$ (see Fig.~\ref{figures_SL}). The original image is piecewise constant and is therefore hardly representative of real medical data. In the modified version, constant surfaces are replaced with a mixture of affine, Gaussian and sinusoidal functions. Since the background is expected to be null in PET, it is set to zero in the phantom image.


	To simulate different acquisition times, the pixel intensities of the original phantom $\u_0$ are multiplied by a constant positive factor $\beta$ (13 values in $[10^{-2},10^2]$). The bigger $\beta$, the higher the number of photon count per pixel and the smaller the relative effect of the noise. To simulate the blur, the $128\times 128$ images are convolved with a normalized Gaussian kernel with standard deviation equal to 1.17 pixel. This PSF is assumed to be ideal, \emph{i.e.}, to be constant over the whole FOV. Finally, each pixel of the image is corrupted by Poisson noise. Algorithm \ref{algo_PD_PET} was used to deblur and denoise the images with TGV regularization (as well as TV with modification of the proximal operators). Parameter $\lambda$ is updated at each of the 20 meta-iterations according to the following rule:
	$$ \lambda^{(l+1)} = \lambda^{(l)} \frac{\text{KL}(\z,\K\hat\u_\lambda)}{M/2},$$
	where the updating factor is called ``KL ratio''. The signal-to-noise ratio (SNR) is used to quantify the quality of the corrupted image ($\text{SNR}_{in}$) and the restored image ($\text{SNR}_{out}$) relative to the original one. Convolution and noise influence the $\text{SNR}_{in}$. SNR between image $\x$ and the original image $\y$ is defined as 
	$$  \text{SNR}(\x,\y) = 20\log_{10}\left(\frac{\|\x\|_2}{\|\x-\y\|_2}\right).$$

	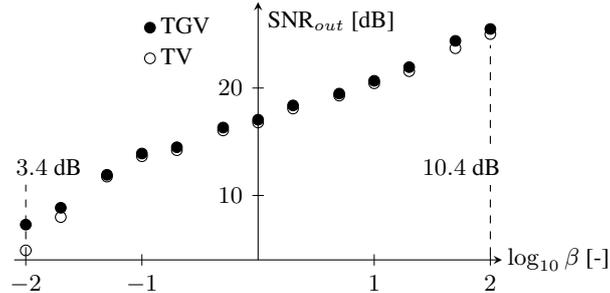
\begin{figure}
\begin{center}
\begin{tikzpicture}
\begin{axis}[
	font = \small,
	height = 5cm,
	width = 8cm,
	scale = 1,
	axis lines = middle,
	xmin = -2.1, xmax = 2.1, 
	ymin = 4, ymax = 28, 
	xlabel = {$\log_{10}{\beta}$ [-]}, ylabel = {SNR$_{out}$ [dB]},
	every x tick/.style={color=black},
	every y tick/.style={color=black},
	every axis legend/.append style={at={(0.25,0.7)},anchor=south west, draw=none,legend cell align=left}, 
	axis labels at tip
	]

\addplot+ [mark options={black}, only marks] file
	{anoise_TGV.dat};
\addlegendentry{TGV}
\addplot+ [mark = o, mark options={black}, only marks] file
	{anoise_TV.dat};
\addlegendentry{TV}

\draw[dashed] (axis cs:-2,0) |- (axis cs:-2,11);
\draw(axis cs:-1.8,11) node[above]{\small{$3.4~\text{dB}$}};
\draw[dashed] (axis cs:2,0) |- (axis cs:2,24);
\draw(axis cs:1.75,11) node[fill=white,above]{\small{$10.4~\text{dB}$}};

\end{axis} 
\end{tikzpicture}
\end{center}
\vskip-4ex
\caption{\small Output SNR as a function of the level of noise $\beta$ for TGV and TV regularizations. Input SNR are indicated for extremal values.}
\label{figure_SNR_out}
\end{figure}
	
\begin{figure}[b!]
\begin{center}
\begin{tikzpicture}
\begin{axis}[
	font = \small,
	height = 5cm,
	width = 8cm,
	scale = 1,
	axis lines = middle,
	xmin = 2, xmax = 21, 
	ymin = 0.9, ymax = 1.7, 
	xlabel = {\# iteration}, ylabel = {KL ratio},
	every x tick/.style={color=black},
	every y tick/.style={color=black},
	every axis/.append style={font=\tiny},
	every axis legend/.append style={at={(1,0.3)},anchor=south east}, 
	xtick = {2,4,...,20},
 	ytick = {1,1.2,1.4},
	axis labels at tip
	]
	
\addplot+ [smooth,no marks, black] table[x expr=\coordindex + 2, y index = 0]
	{KL_plusstd_TGV_DC.dat};
\addplot+ [smooth,mark options={black},black, thick] table[x expr=\coordindex + 2, y index = 0]
	{KL_mean_TGV_DC.dat};
\addplot+ [smooth,no marks, black] table[x expr=\coordindex + 2, y index = 0]
	{KL_minusstd_TGV_DC.dat};
	
\draw[dashed] (axis cs:2,1) |- (axis cs:22,1);
\end{axis} 
\end{tikzpicture}
\end{center}
\vskip-4ex
\caption{\small KL ratio evolution as a function of the meta-iteration count. The thick black curve indicates the mean of 13 realizations (different noise levels) and the thin curve indicates the standard deviation.}
\label{figure_lambda}
\end{figure}
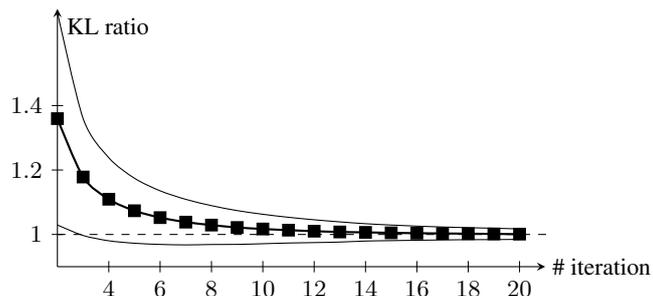

	\textbf{Results.} A visual comparison between TGV and TV regularizations is presented on Fig.~\ref{figures_SL} for two levels of noise: $\beta = 1$ and $\beta = 0.1$. The impact of the noise level (or similarly, the acquisition time) on the SNR is illustrated in Fig.~\ref{figure_SNR_out} for both TGV and TV regularizations of the same corrupted images. The input SNR is indicated for extremal values. Finally, Fig.~\ref{figure_lambda} shows the evolution of the KL ratio as a function of the meta-iterations for automatic selection of parameter $\lambda$.
		

\textbf{Patient images.}
\label{subsubsec:exp_method}
PET images of patients with pharyngolaryngeal squamous cell carcinoma were acquired on a Siemens Ecat HR scanner for a previous study \cite{Daisne2004}. The size of the image is $128\times 128 \times 47$ and the voxel size is $2.2\times 2.2 \times 3.12~mm^3$. The algorithm was applied on axial slices. The PSF was measured experimentally with a line source perpendicular to the axial slices in diffusing material (water). In first approximation, the PSF at $100~mm$ from the FOV center was Gaussian, isotropic, with $6~mm$ of FWHM. Algorithm \ref{algo_PD_PET} was used to deblur and denoise the images. The value of the optimal regularization parameter was set to $\lambda=5$ (see Sec.~\ref{sec:discussion}). 

\textbf{Results.} Since there is no original image available, main results consist of a visual comparison between TGV and TV regularizations (see Fig.~\ref{figures_PET}).

\section{Discussion}
\label{sec:discussion}

	\textbf{Synthetic images.} Results on Fig.~\ref{figures_SL} show that both methods remove the noise and preserve edges. Some staircasing artifacts affect TV restored images since the original image is not piecewise constant. This cartoon-like appearance removes details like the small white disks. As expected, TGV restored images keep the smoothness property of the original image. The value of parameter $\lambda$ depends on the image dynamics (bigger if higher intensity) and is similar for the two regularization methods. Distinction between TGV and TV is not so obvious on the graph of the SNR evolution. In average, $\text{SNR}_{out}$ of TGV is $0.5~\text{dB}$ above TV's and is mostly due to the contribution of small $\beta$ ($2~\text{dB}$ for $\beta=10^{-2}$).
The updating rule for parameter $\lambda$ gives convincing results since the KL ratio converges to 1 and the standard deviation to 0 (average on 13 realizations). These results validate the automatic selection rule of $\lambda$ in case of Poisson noise. Regarding Fig.~\ref{figure_lambda}, 10 iterations seem to be enough for accurate estimation of $\lambda$. 

\begin{figure}	
	\captionsetup[subfigure]{labelformat=empty}
	\centering
	\subfloat{\includegraphics[width=\sizeFig]{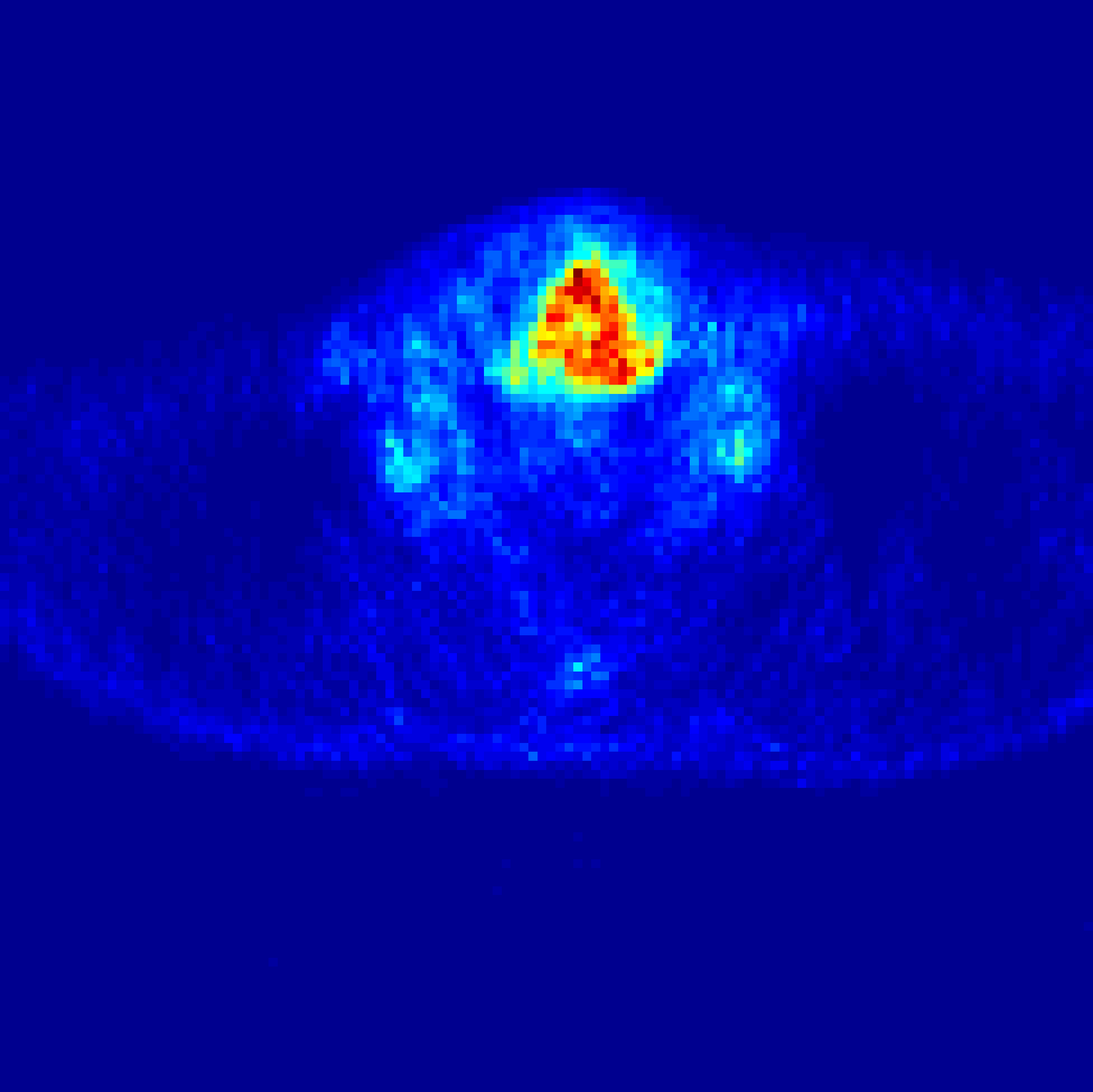}} \hspace{\hspaceFig}
	\subfloat{\includegraphics[width=\sizeFig]{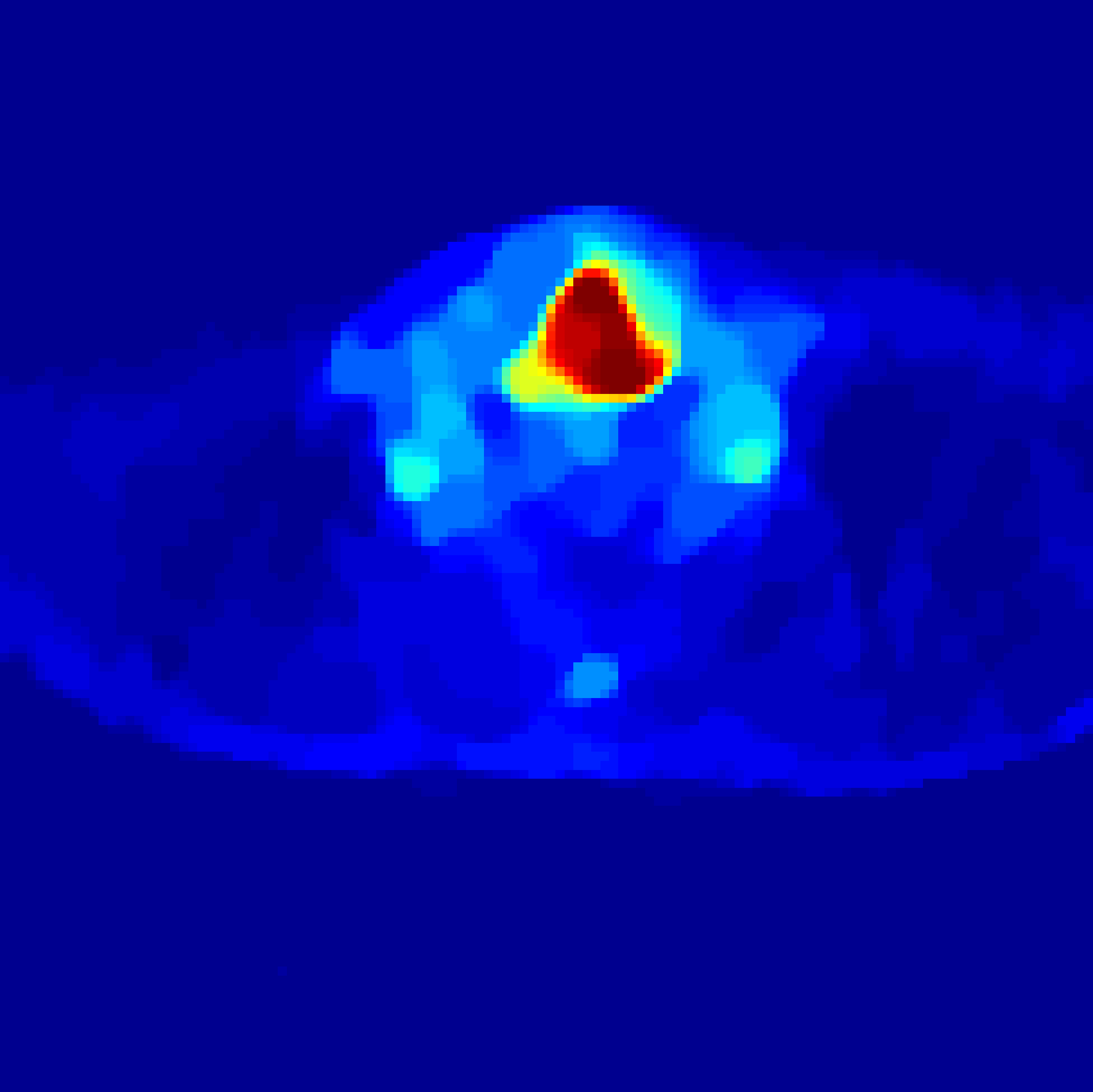}} \hspace{\hspaceFig}
	\subfloat{\includegraphics[width=\sizeFig]{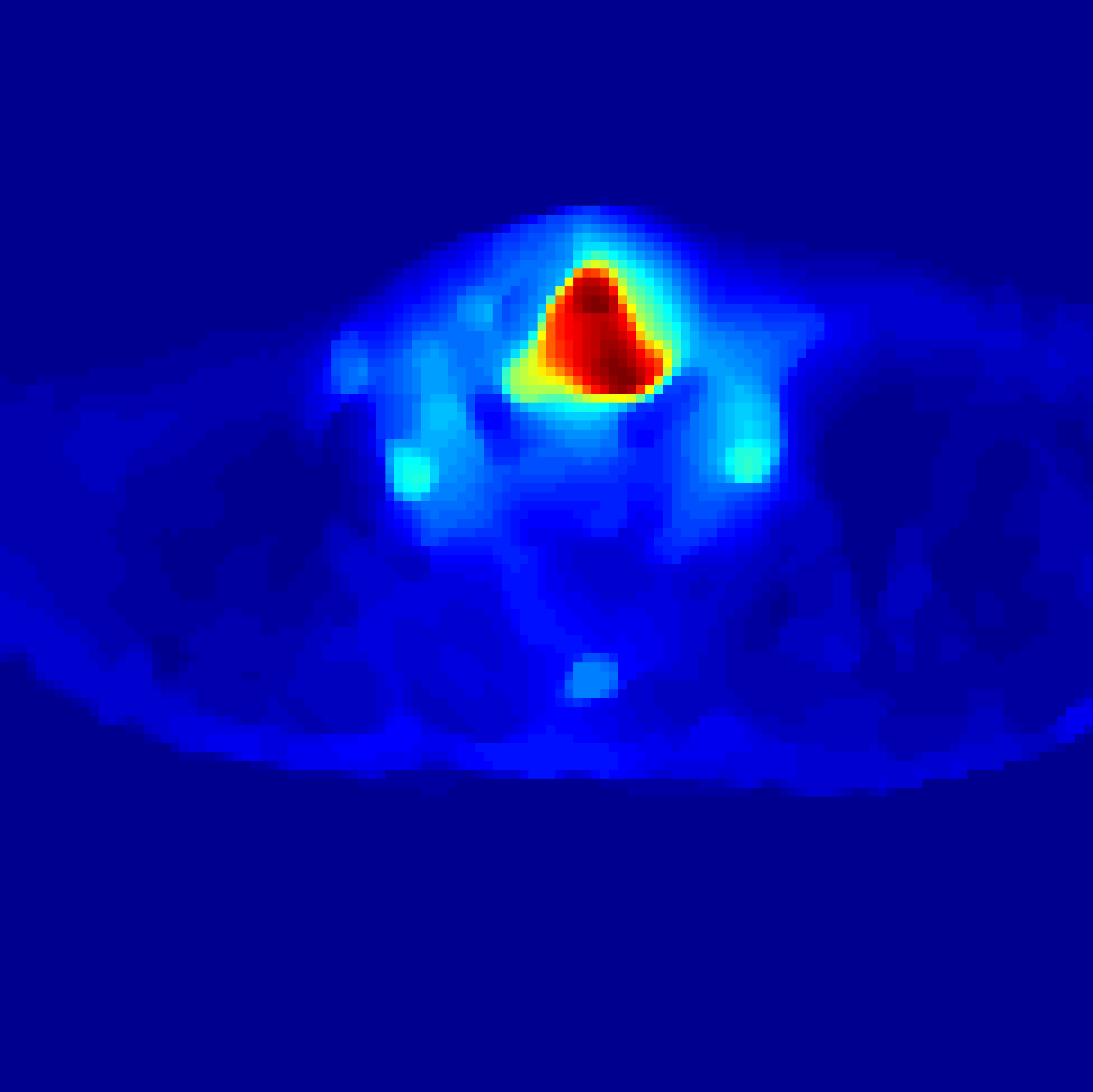}}\\[1mm]
	\subfloat{\includegraphics[width=\sizeFig]{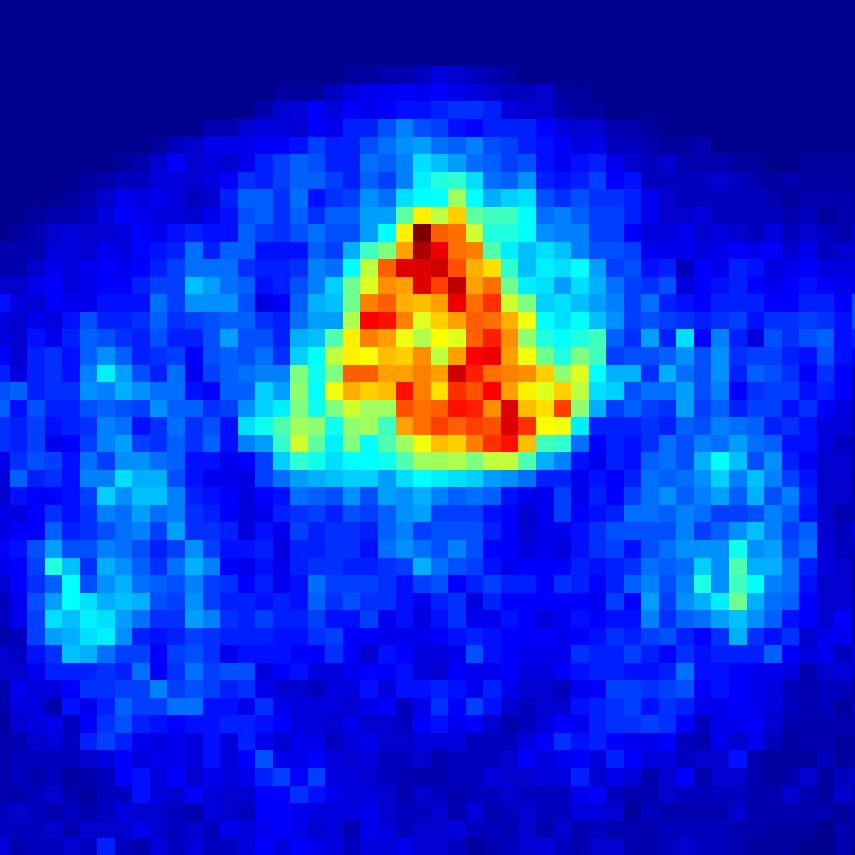}} \hspace{\hspaceFig}
	\subfloat{\includegraphics[width=\sizeFig]{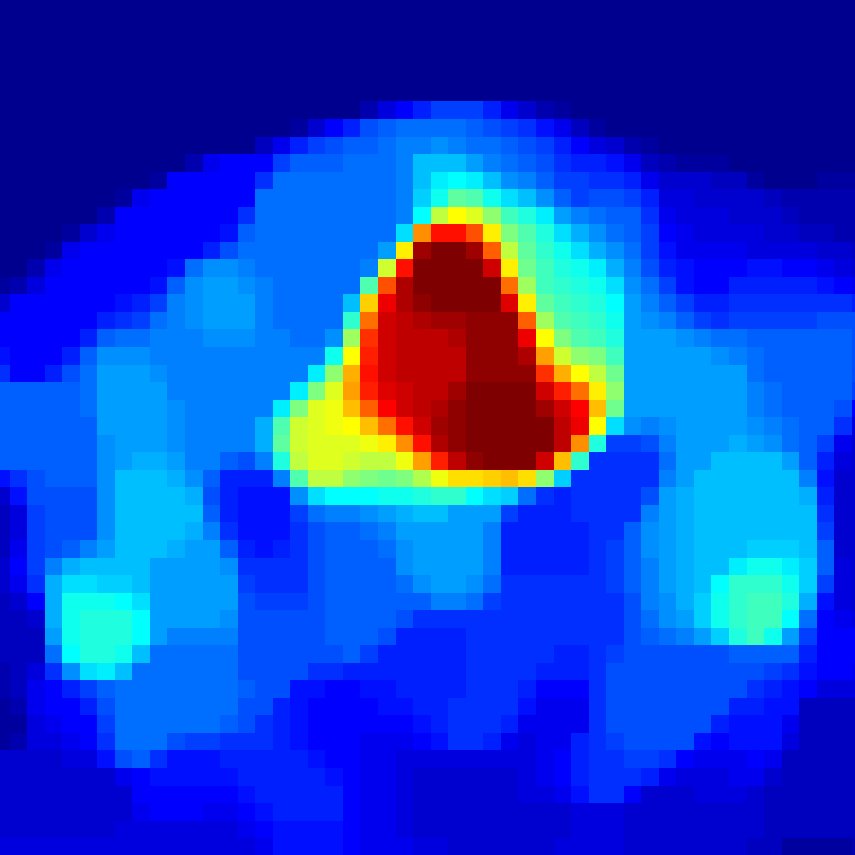}} \hspace{\hspaceFig}
	\subfloat{\includegraphics[width=\sizeFig]{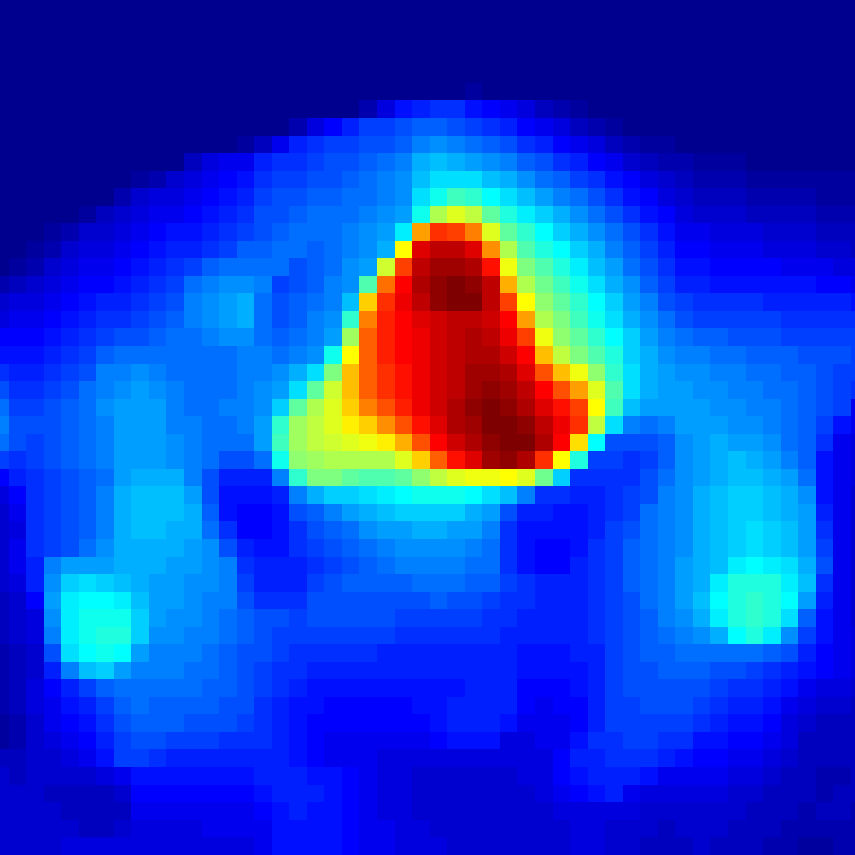}}
	\caption{\small Results on real medical data (full and zoom). From left to right: corrupted PET image, TV and TGV restorations with $\lambda = 5$.}
	\label{figures_PET}
\end{figure}

\textbf{Patient images.} The staircasing artifacts observed in synthetic image with TV restoration are also visible on real medical images (Fig.~\ref{figures_PET}). Since data are not corrupted with a pure Poisson noise (see Sec.~\ref{sec:intro}), the updating rule for regularization parameter $\lambda$ cannot be applied. The choice of the value of $\lambda$ is highly subjective and depends on the application as mentionned in \cite{Anthoine2012}.


\section{Conclusions and future challenges}	
\label{sec:conclusion}
This work presents a new method for deconvolution of PET images based on TGV regularization. The use of TGV instead of TV as a prior is more adapted to real medical images for which the piecewise-constant assumption is not verified. In case of pure Poisson noise, we validate an updating rule for parameter $\lambda$. Further work will investigate the nature of the noise after reconstruction (with Monte-Carlo simulations) to improve the formulation of the inverse problem as well as the updating rule of $\lambda$.
	

\bibliographystyle{IEEEbib}
\bibliography{library}

\end{document}